\documentclass[10pt,twocolumn,letterpaper]{article}

\usepackage[pagenumbers]{cvpr} % To force page numbers, e.g. for an arXiv version

\usepackage[dvipsnames]{xcolor}

\usepackage{tabularx, multirow}
\newcolumntype{Y}{>{\centering\arraybackslash}X}

\definecolor{cvprblue}{rgb}{0.21,0.49,0.74}
\usepackage[pagebackref,breaklinks,colorlinks,citecolor=cvprblue]{hyperref}

\def\paperID{*****} % *** Enter the Paper ID here
\def\confName{CIS700}
\def\confYear{2024}

\title{Multi-Agent VQA: Exploring Multi-Agent Foundation Models in \\ Zero-Shot Visual Question Answering}

\author{Bowen Jiang, Zhijun Zhuang, Shreyas S. Shivakumar, Dan Roth, Camillo J. Taylor\\
GRASP Lab, University of Pennsylvania\\
Philadelphia, PA, 19104, USA\\
{\tt\small \{bwjiang, zhijunz, sshreyas, danroth, cjtaylor\}@seas.upenn.edu}
}

\begin{document}
\maketitle
\begin{abstract}
% This work studies the Visual Question Answering (VQA) problem from three key innovations. First, we explore the zero-shot capabilities of foundation models in VQA, proposing Multi-Agent VQA, an adaptive multi-agent system to overcome their limitations in object detection and counting. Unlike existing approaches, we study the potential of foundation models without fine-tuning, making the system more practical in the open world. Second, we introduce a new benchmark, Zero-Shot VQA, where our method outperforms other zero-shot algorithms across all datasets. It also shows significant improvement in utilizing multi-agents compared to baselines. Lastly, this preliminary work highlights some challenges and interesting failure cases,
% % when applying foundation models in VQA tasks, 
% offering new directions for future research. 

This work explores the zero-shot capabilities of foundation models in Visual Question Answering (VQA) tasks. We propose an adaptive multi-agent system, named Multi-Agent VQA, to overcome the limitations of foundation models in object detection and counting by using specialized agents as tools. Unlike existing approaches, our study focuses on the system's performance without fine-tuning it on specific VQA datasets, making it more practical and robust in the open world. We present preliminary experimental results under zero-shot scenarios and highlight some failure cases, offering new directions for future research. 

\end{abstract}    
\section{Introduction}
\label{sec:intro}
Recently we have witnessed a rapid emergence of multi-modal foundation models~\cite{bommasani2021opportunities, OpenAI2023GPT4V, radford2019language}, that seamlessly bridge vision and language understanding tasks.
% Recent advances in foundation models have witnessed a rapid emergence of multi-model representation learning, which seamlessly bridges vision and language
% , the two most fundamental modalities, for joint understanding.
% in the real world. 
This fusion allows 
% language models to finally see the world, while enabling 
vision systems to leverage the versatility of natural language, thereby extending their understanding and reasoning capabilities to an unprecedented level. 

Visual Question Answering (VQA)~\cite{goyal2017making} serves as a suitable problem in testing foundation models on complicated vision-language understanding. 
% The task is intrinsically challenging because users may inquire about any aspect of an image in unrestricted question formats and sometimes require the model to apply common sense reasoning that was impossible before the era of large language models (LLM). 
% Due to the challenges it poses, VQA remains a hard problem worth attention to push the frontier of foundation models.
Despite their popularity, the zero-shot capabilities of foundation models in the domain of VQA remain largely unexplored. Almost all pretrained large vision-language models (LVLM) in the VQA literature require fine-tuning on specified VQA datasets with a very limited vocabulary to achieve state-of-the-art performance~\cite{chen2022pali, wang2022image, wang2023one}. This practice, while effective, overlooks the actual potential of foundation models and
% to adapt to new tasks and
% through in-context learning~\cite{dong2022survey} and multi-agents~\cite{wu2023autogen}, as well as 
limits their open-world usage beyond their fine-tuned datasets. The generalization ability of foundation models already demonstrated in many multi-modal tasks suggests that an exploration of their zero-shot VQA performance would unveil new dimensions of their abilities. 

This work uncovers some of the challenges these models might face. While foundation models are typically pretrained with images and corresponding textual descriptions, they may not have been specifically pretrained to interpret underlying graphical structures in images. As a result, they often fall short in VQA when the question details specific object attributes and relationships that form local scene graphs~\cite{zhao2023less}, or counting the number of objects in images. 
% In scenarios where a foundation model overlooks key objects in the global scene, it will inevitably fail to answer the question accurately. 

Tools~\cite{schick2024toolformer} are specialized agents in their own fields. Instead of fine-tuning the foundation models to overcome their limitations, we harness specialized models for object detection and counting as tools within our multi-agent system~\cite{wu2023autogen}.
% Thanks to the idea of using tools and multi-agents~\cite{schick2024toolformer, wu2023autogen}, we aim to tackle such limitations by integrating specialized agents for object detection and counting as tools within our multi-agent system.
Specifically, when an LVLM fails to detect an object in the scene, an object detector like Grounded Segment Everything~\cite{ren2024grounded} can help. Likewise, for questions that require counting objects, a model specifically trained for the counting task~\cite{jiang2023clip} can fulfill this role. This collaborative framework enables a more flexible response to the diverse challenges in VQA tasks, effectively exploiting the full capability of foundation models without additional training.

% To summarize our contributions:

\begin{figure*}[t]
  \centering
    \includegraphics[width=0.78\linewidth]{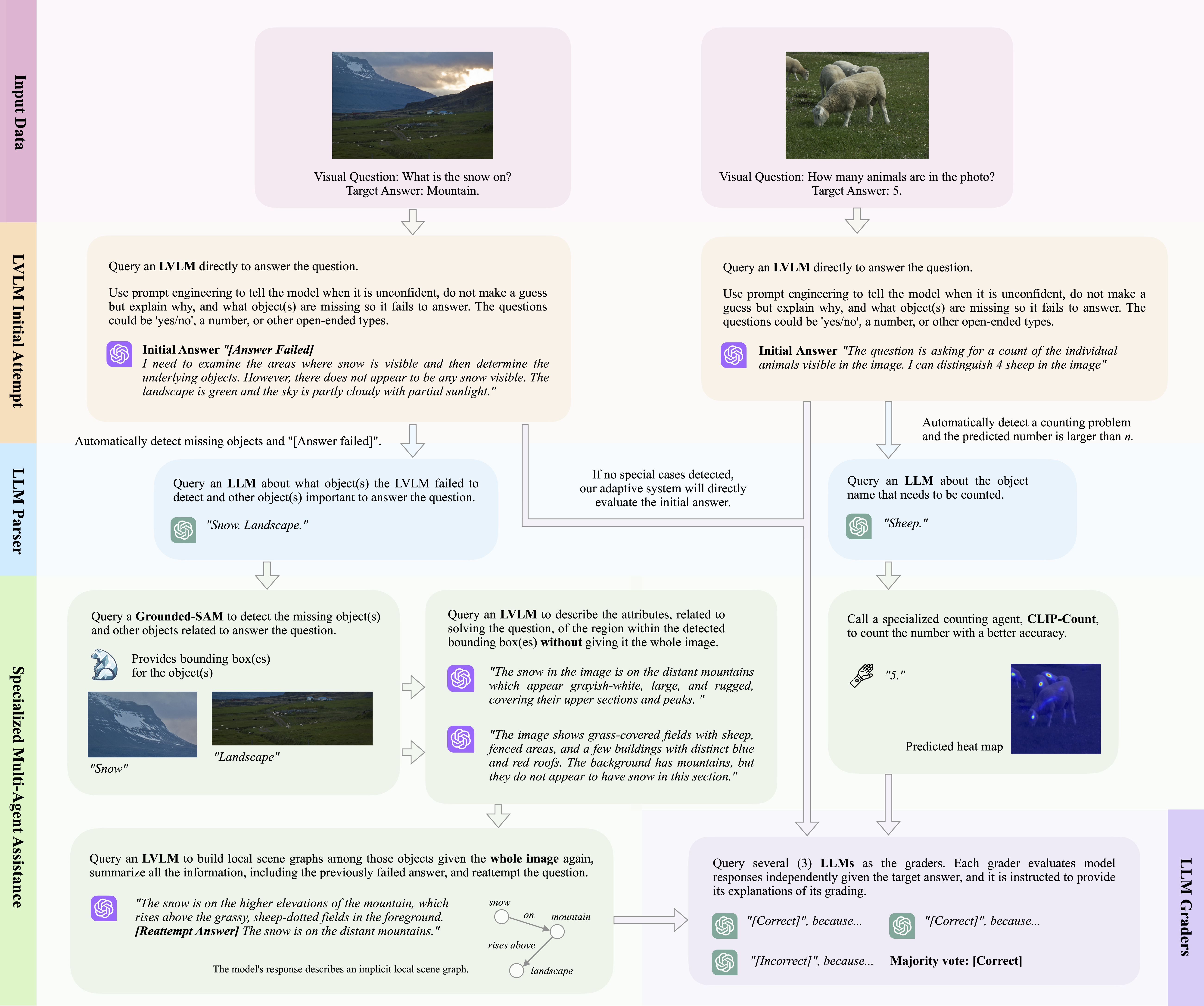}
    \caption{Overview of the adaptive Multi-Agent VQA System. The process begins with an LVLM attempting to answer a visual question directly. An LLM parsing agent automatically detects challenging cases and calls specialized agents, including object-detection and counting models as tools. The LVLM would reattempt the question, with the response assessed by LLM-based graders for a majority vote.}
    \label{fig:pipeline}
    \vspace{-3mm}
\end{figure*}

\section{Methods}
\label{sec:method}

This section describes the pipeline of our adaptive Multi-Agent VQA system, shown in Figure~\ref{fig:pipeline}. 
% It is a general pipeline that works for different VQA tasks, where 
Different agents guide the system to analyze shortcomings, fill in missing information, and discover the final answer step-by-step. We use GPT-4V~\cite{OpenAI2023GPT4V} as our LVLM and GPT-3.5~\cite{radford2019language} as our LLM, but the system can be adapted to other foundation models.

\subsection{Initial attempt}
% We acknowledge the inherent strength of large vision-language models (LVLM) in directly responding to a variety of visual questions. Therefore, 
We introduce an adaptive pipeline that allows an LVLM to answer a given question directly, which can optimize the average inference time. We carefully craft prompts to introduce the problem and guide the LVLM to avoid overconfidence.
% and engage in chain-of-though reasoning~\cite{wei2022chain}. 
If the LVLM thinks it cannot produce the answer because it has missed key objects in the image, it is instructed to say so explicitly using the following special tokens ``[Answer Failed]".
% , and list the missing objects. 
Otherwise, the system will 
% proceed with ``[Answer]", 
bypass the multi-agent modules to avoid unecessary additional computation.

\subsection{Reattempt by adaptively calling multi-agents}
\subsubsection{LLM parsing agent}
An LLM-based parsing agent aims to automatically detect special cases in the initial response and select appropriate agents for assistance. 
Specifically, whenever it detects the special tokens ``[Answer Failed]", it will extract the object names that the LVLM thought were missing. We also realize that LVLMs may struggle with counting objects, especially when the number $n$ is large. The LLM parsing agent will detect whether the question refers to a counting problem and extract the named objects that need to be counted.
\vspace{-7.5mm}

% An LLM-based parsing agent is implemented to detect if special tokens ``[Answer Failed]" exist in the previous response. Upon detection, the parsing agent extracts the names of the objects that the LVLM thought as missing.
% % , and formats them into a list like ``Object 1. Object 2." accessible for a subsequent object detection agent to fill in the gaps.

% We realize that foundation models have a limitation in accurately counting objects especially when the number $n$ is larger than three or four.
% % To overcome this limitation without fine-tuning the LVLM, 
% Therefore, the parsing agent is also responsible for detecting whether the question refers to a counting problem and if the initially predicted number is larger than $n$. If so, it needs to extract the object names to be sent to a specialized object-counting agent for reassessment.

\subsubsection{Object detection agent}
\vspace{-0.5mm}
Following the identification of missing objects, our system employs a pretrained object-detection agent to localize them in the visual scene. We rely on the Grounded Segment Everything model~\cite{ren2024grounded} as the tool, which takes object names as inputs and outputs masks and bounding boxes. Unlike general LVLMs, it is specifically trained in object detection tasks is better at recognizing and localizing small or non-obvious objects that the initial LVLM might overlook. 
\vspace{-1mm}

\subsubsection{LVLM object description agent}
This step bridges the object-detection agent and the original LVLM, focusing on the detailed analysis of important objects detected in the previous step. We crop the original image based on the predicted bounding boxes and feed each region to a separate instance of the LVLM in parallel. These LVLMs have identical architecture and weights, tasked with describing attributes related to answering the question for each object. For example, one LVLM object-description agent describes the snow in Figure~\ref{fig:pipeline} as covering a \textit{``grayish-white distant mountain"}, pointing out the key information that can finally solve the question \textit{``what is the snow on"} successfully. By narrowing each LVLM's field of view to just the cropped region within the bounding box, we eliminate the complexity of scanning the entire visual scene to localize relevant objects and their visual details. 
\vspace{-1mm}

\subsubsection{LVLM reattempts the question}
Finally, we revisit the original LVLM by giving it a comprehensive set of inputs: the original image and question, the initially unsuccessful attempt and why it failed, and the description of the detected objects that were previously missing. The model now is expected to have enough context to construct an implicit local scene graph expressed in natural language, figuring out relationships among these objects that are relevant to answering the question. Take the example in Figure~\ref{fig:pipeline} again; the model would now understand that the snow is only located on the mountain but not on the broader landscape. Such an implicit graphical structure allows the LVLM to weave together the various pieces of information into a whole and reattempt the visual question to provide a more accurate and comprehensive response.
\vspace{-2mm}

\subsubsection{Object counting agent}
In scenarios where the LLM parsing agent recognizes a counting problem and the initially predicted number $n$ is greater than $3$, the system will call a specialized counting agent, CLIP-Count~\cite{jiang2023clip}, for help. Guided by text prompts, CLIP-Count generates a density map for open-vocabulary objects in a zero-shot manner and achieves better counting accuracy. Its answer will serve as the final answer.

\subsection{LLM answer grading agent}
Unlike existing approaches that fine-tune models on given datasets~\cite{chen2022pali, wang2022image, wang2023one}, we no longer expect the model to replicate exact dataset annotations~\cite{goyal2017making}. By embracing zero-shot learning, we avoid training foundation models to fit a limited vocabulary and the inevitable human annotation bias. To this end, we introduce LLM-based graders as a novel component of our framework for open-ended evaluation, mimicking humans that allow diverse phrasing and additional information not mentioned in the annotation. We collect assessments from three individual grading agents and take the majority vote to provide a more robust evaluation.

\section{Experiments}
\subsection{Datasets}
We evaluate our method on the widely adopted VQA-v2~\cite{goyal2017making} and GQA~\cite{hudson2019gqa} datasets. Due to the costs and time requirements of GPT-4V API~\cite{OpenAI2023GPT4V},  we have to use a subset of the data to evaluate the performance - a typical drawback in large foundation models. For GQA, we take the same $1000$ validation samples used in~\cite{zhao2023less} for testing. VQA-v2 comprises ``yes/no", ``number", and other question types. However, its test set is not publicly available and requires exact matches of the answers, making our LLM-based graders inapplicable. We instead adopt the VQA-v2 rest-val dataset, the validation dataset in \cite{wang2022image, bao2022vlmo} that was never used for training. It contains $5228$ unique image-question pairs.

\subsection{Results and Zero-Shot VQA benchmark}
We split Table~\ref{tab:vqav2} into fine-tuned and zero-shot sections for fair comparisons. 
% When current state-of-the-art methods, 
We run BEiT-3~\cite{wang2022image} and VLMo~\cite{bao2022vlmo}, representing the current state of the art. When the BEiT-3~\cite{wang2022image} or VLMo model~\cite{bao2022vlmo} fine-tuned on VQA-v2 training dataset is evaluated on VQA-v2 rest-val, namely \textit{BEiT3-large-vqa-v2} or \textit{VLMo-large-vqa-v2} in Table~\ref{tab:vqav2}, it keeps its advantage. 

However, versions of BEiT-3 or VLMo that have \emph{not} been fine-tuned on VQA-v2, namely \textit{BEiT3-large-in-domain} or \textit{VLMo-large-coco} in Table~\ref{tab:vqav2}, achieve almost zero accuracies on VQA-v2. Table~\ref{tab:gqa} also shows significant drops in their performance on GQA in zero-shot, a dataset different from the one, namely VQA-v2, they are fine-tuned on. \textbf{These declines highlight a significant limitation of existing VQA models:} Despite their advancements, models like BEiT-3 and VLMo depend heavily on dataset-specific fine-tuning with a low zero-shot generalization ability. Most existing works share similar designs~\cite{chen2022pali, wang2022image, wang2023one}, underscoring the unique value of our zero-shot solution.
\vspace{-1mm}

\begin{table}[h]
  \centering
  \small
  \caption{Results on VQA-v2 rest-val~\cite{goyal2017making} dataset.}
  \vspace{-3mm}
  \begin{tabular*}{\columnwidth}{@{\extracolsep{\fill}} ccc} % Spread columns evenly
    \toprule
    Weights & Methods & Accuracy \\
    \midrule
    \multirow{2}{*}{Fine-tuned} & BEiT3-large-vqa-v2~\cite{wang2022image} & \textbf{88.33} \\
    & VLMo-large-vqa-v2~\cite{bao2022vlmo} & 83.36 \\
    \midrule
    \multirow{3}{*}{Zero-shot} 
                            & BEiT3-large-indomain~\cite{wang2022image} & 0.01 \\
                            & VLMo-large-coco~\cite{bao2022vlmo}  & 0.00 \\
                            & Multi-Agent VQA (ours) & \textbf{78.02} \\
    \bottomrule
  \end{tabular*}
  \vspace{-6mm}
  \label{tab:vqav2}
\end{table}
\begin{table}[h]
  \centering
  \small
  \caption{Results on GQA-val~\cite{hudson2019gqa, zhao2023less} subset.}
  \vspace{-3mm}
  \begin{tabular*}{\columnwidth}{@{\extracolsep{\fill}} ccc} % Spread columns evenly
    \toprule
    Weights & Methods & Accuracy \\
    \midrule
    \multirow{5}{*}{Zero-shot}
                            & BEiT3-large-vqa-v2~\cite{wang2022image} & 64.67 \\
                            & VLMo-large-vqa-v2~\cite{bao2022vlmo} & 0.00 \\
                            % & LessIsMore-global & 55.40 \\
                            & BLIP2-flan-t5-xl~\cite{li2023blip} & 50.40 \\
                            & LessIsMore-local~\cite{zhao2023less} & 58.30 \\
                            & Multi-Agent VQA (ours) & \textbf{79.70} \\
    \bottomrule
  \end{tabular*}
  \vspace{-3mm}
  \label{tab:gqa}
\end{table}

\subsection{Ablation study}
W assess the impact of removing detailed chain-of-thought (CoT) reasoning~\cite{wei2022chain}, the CLIP-Count object-counting agent~\cite{jiang2023clip}, and the multi-agent pipeline in Table~\ref{tab:ablation}. Keeping only basic prompt instructions leads to reduced performance in all question types. The removal of CLIP-Count forces the LVLM to answer all counting questions itself, resulting in an accuracy drop on this type of question, labeled as ``num" in Table~\ref{tab:ablation}. The most profound impact happens by removing the proposed multi-agent pipeline, where the system has to rely solely on a single LVLM for VQA tasks, here the accuracy decreases by nearly $10$ percent.

\begin{table}[h]
  \centering
  \small
  \caption{Ablation study}
  \vspace{-3mm}
  \begin{tabular*}{\columnwidth}{@{\extracolsep{\fill}} ccc} % Spread columns evenly
    \toprule
    Dataset & Ablation & Acc (yes/no, num, other) \\
    \midrule
    \multirow{4}{*}{VQA-v2} & w/o detailed CoT~\cite{wei2022chain} & 74.81 (81.20, 57.01, 74.54) \\
                            % & w/o detailed CoT \& multi-agent & 53.62 \\
                            & w/o CLIP-Count~\cite{jiang2023clip} & 76.47 (84.26, 52.81, 76.50) \\
                            % & w/o CLIP-Count \& multi-agent & 66.58 \\
                            & w/o multi-agent & 69.20 (79.05, 53.43, 65.21) \\
                            & final & \textbf{78.02} (\textbf{84.82}, \textbf{60.63}, \textbf{77.83}) \\
    \midrule
    \multirow{2}{*}{GQA} & w/o multi-agent & 68.50 \\
                         & final & \textbf{79.70} \\
    \bottomrule
  \end{tabular*}
  \vspace{-3mm}
  \label{tab:ablation}
\end{table}

\subsection{Limitations and failure examples}
\vspace{-0.5mm}
The performance in object counting tasks is bounded by the counting agent which does not always work well, and there are currently few methods that support zero-shot object counting~\cite{jiang2023clip}. The reliance on API calls for LLM and LVLM agents constrains the inference speed of our model and the outputs are not entirely deterministic.

\begin{figure}[h]
  \centering
    \includegraphics[width=1\linewidth]{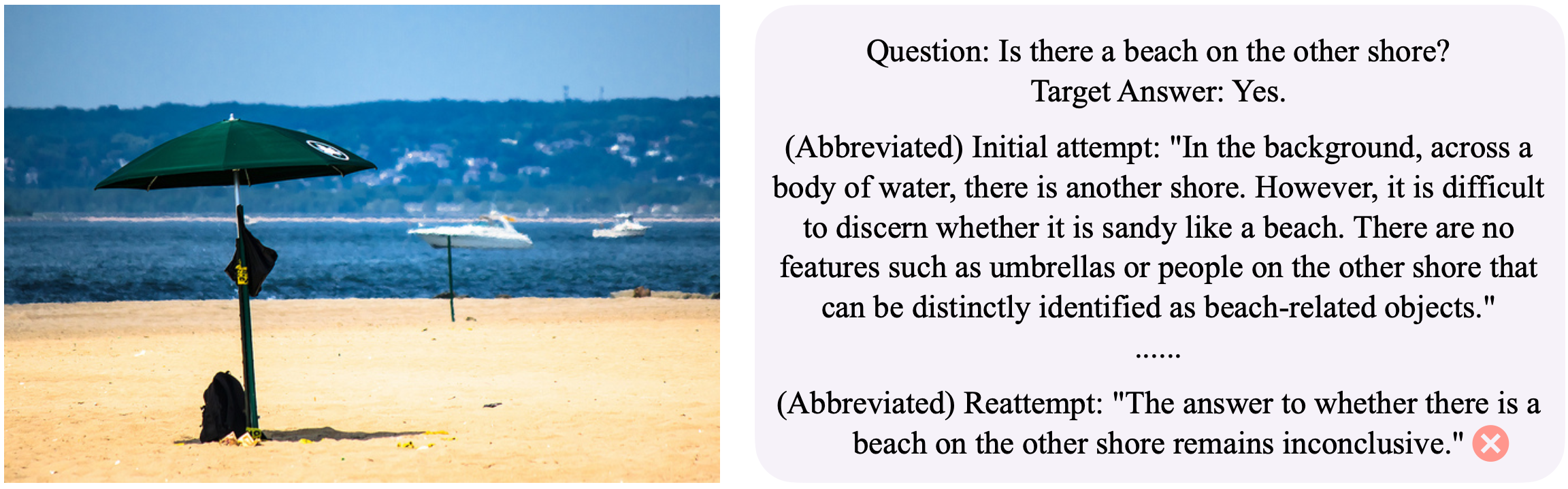}
    \vspace{-5mm}
    \caption{An example of the failure case.}
    \label{fig:fail}
    \vspace{-6mm}
\end{figure}

Failure cases often happen when the foundation model overthinks the questions and provides more exhaustive answers than necessary. Take Figure~\ref{fig:fail} as an example; the model accurately deduces the presence of a shore on the opposite side but unnecessarily tries to figure out if the shore is sandy and, therefore, a beach. It also calls the object-detection agent to spot umbrellas, a typical sign of a beach, but it wastes time in analyzing the umbrella on the current side of the shore.

\label{sec:experiment}

% \section{Conclusion}
% \label{sec:conclusion}
% Our work explores foundation models on VQA tasks and employs an adaptive multi-agent system to address limitations by using tools. This approach marks a departure from existing works that heavily rely on dataset-specific fine-tuning, showcasing a competitive zero-shot performance.
% % , and we introduce a Zero-Shot VQA benchmark to encourage further research in this direction. 
% Our findings represent preliminary work, and we expect to conduct more comprehensive experiments and zero-shot benchmarks in the near future.

\section{Future work}
\label{sec:conclusion}
Our findings represent preliminary work on the zero-shot VQA capabilities of foundation models. In the near future, we will employ different foundation models and specialized tools, discuss detailed prompt engineering and chain-of-thought reasoning, and present a more comprehensive zero-shot VQA benchmark in the open world.\hspace{-3cm}

{
    \small
    \bibliographystyle{ieeenat_fullname}
    \bibliography{main}

\begin{thebibliography}{16}
\providecommand{\natexlab}[1]{#1}
\providecommand{\url}[1]{\texttt{#1}}
\expandafter\ifx\csname urlstyle\endcsname\relax
  \providecommand{\doi}[1]{doi: #1}\else
  \providecommand{\doi}{doi: \begingroup \urlstyle{rm}\Url}\fi

\bibitem[Bao et~al.(2022)Bao, Wang, Dong, Liu, Mohammed, Aggarwal, Som, Piao, and Wei]{bao2022vlmo}
Hangbo Bao, Wenhui Wang, Li Dong, Qiang Liu, Owais~Khan Mohammed, Kriti Aggarwal, Subhojit Som, Songhao Piao, and Furu Wei.
\newblock Vlmo: Unified vision-language pre-training with mixture-of-modality-experts.
\newblock \emph{Advances in Neural Information Processing Systems}, 35:\penalty0 32897--32912, 2022.

\bibitem[Bommasani et~al.(2021)Bommasani, Hudson, Adeli, Altman, Arora, von Arx, Bernstein, Bohg, Bosselut, Brunskill, et~al.]{bommasani2021opportunities}
Rishi Bommasani, Drew~A Hudson, Ehsan Adeli, Russ Altman, Simran Arora, Sydney von Arx, Michael~S Bernstein, Jeannette Bohg, Antoine Bosselut, Emma Brunskill, et~al.
\newblock On the opportunities and risks of foundation models.
\newblock \emph{arXiv preprint arXiv:2108.07258}, 2021.

\bibitem[Chen et~al.(2022)Chen, Wang, Changpinyo, Piergiovanni, Padlewski, Salz, Goodman, Grycner, Mustafa, Beyer, et~al.]{chen2022pali}
Xi Chen, Xiao Wang, Soravit Changpinyo, AJ Piergiovanni, Piotr Padlewski, Daniel Salz, Sebastian Goodman, Adam Grycner, Basil Mustafa, Lucas Beyer, et~al.
\newblock Pali: A jointly-scaled multilingual language-image model.
\newblock \emph{arXiv preprint arXiv:2209.06794}, 2022.

\bibitem[Goyal et~al.(2017)Goyal, Khot, Summers-Stay, Batra, and Parikh]{goyal2017making}
Yash Goyal, Tejas Khot, Douglas Summers-Stay, Dhruv Batra, and Devi Parikh.
\newblock Making the v in vqa matter: Elevating the role of image understanding in visual question answering.
\newblock In \emph{Proceedings of the IEEE conference on computer vision and pattern recognition}, pages 6904--6913, 2017.

\bibitem[Hudson and Manning(2019)]{hudson2019gqa}
Drew~A Hudson and Christopher~D Manning.
\newblock Gqa: A new dataset for real-world visual reasoning and compositional question answering.
\newblock In \emph{Proceedings of the IEEE/CVF conference on computer vision and pattern recognition}, pages 6700--6709, 2019.

\bibitem[Jiang et~al.(2023)Jiang, Liu, and Chen]{jiang2023clip}
Ruixiang Jiang, Lingbo Liu, and Changwen Chen.
\newblock Clip-count: Towards text-guided zero-shot object counting.
\newblock \emph{arXiv preprint arXiv:2305.07304}, 2023.

\bibitem[Li et~al.(2023)Li, Li, Savarese, and Hoi]{li2023blip}
Junnan Li, Dongxu Li, Silvio Savarese, and Steven Hoi.
\newblock Blip-2: Bootstrapping language-image pre-training with frozen image encoders and large language models.
\newblock In \emph{International conference on machine learning}, pages 19730--19742. PMLR, 2023.

\bibitem[OpenAI(2023)]{OpenAI2023GPT4V}
OpenAI.
\newblock Gpt-4v(ision) system card.
\newblock \emph{cdn.openai.com}, 2023.

\bibitem[Radford et~al.(2019)Radford, Wu, Child, Luan, Amodei, Sutskever, et~al.]{radford2019language}
Alec Radford, Jeffrey Wu, Rewon Child, David Luan, Dario Amodei, Ilya Sutskever, et~al.
\newblock Language models are unsupervised multitask learners.
\newblock \emph{OpenAI blog}, 1\penalty0 (8):\penalty0 9, 2019.

\bibitem[Ren et~al.(2024)Ren, Liu, Zeng, Lin, Li, Cao, Chen, Huang, Chen, Yan, Zeng, Zhang, Li, Yang, Li, Jiang, and Zhang]{ren2024grounded}
Tianhe Ren, Shilong Liu, Ailing Zeng, Jing Lin, Kunchang Li, He Cao, Jiayu Chen, Xinyu Huang, Yukang Chen, Feng Yan, Zhaoyang Zeng, Hao Zhang, Feng Li, Jie Yang, Hongyang Li, Qing Jiang, and Lei Zhang.
\newblock Grounded sam: Assembling open-world models for diverse visual tasks, 2024.

\bibitem[Schick et~al.(2024)Schick, Dwivedi-Yu, Dess{\`\i}, Raileanu, Lomeli, Hambro, Zettlemoyer, Cancedda, and Scialom]{schick2024toolformer}
Timo Schick, Jane Dwivedi-Yu, Roberto Dess{\`\i}, Roberta Raileanu, Maria Lomeli, Eric Hambro, Luke Zettlemoyer, Nicola Cancedda, and Thomas Scialom.
\newblock Toolformer: Language models can teach themselves to use tools.
\newblock \emph{Advances in Neural Information Processing Systems}, 36, 2024.

\bibitem[Wang et~al.(2023)Wang, Wang, Lin, Bai, Zhou, Zhou, Wang, and Zhou]{wang2023one}
Peng Wang, Shijie Wang, Junyang Lin, Shuai Bai, Xiaohuan Zhou, Jingren Zhou, Xinggang Wang, and Chang Zhou.
\newblock One-peace: Exploring one general representation model toward unlimited modalities.
\newblock \emph{arXiv preprint arXiv:2305.11172}, 2023.

\bibitem[Wang et~al.(2022)Wang, Bao, Dong, Bjorck, Peng, Liu, Aggarwal, Mohammed, Singhal, Som, et~al.]{wang2022image}
Wenhui Wang, Hangbo Bao, Li Dong, Johan Bjorck, Zhiliang Peng, Qiang Liu, Kriti Aggarwal, Owais~Khan Mohammed, Saksham Singhal, Subhojit Som, et~al.
\newblock Image as a foreign language: Beit pretraining for all vision and vision-language tasks.
\newblock \emph{arXiv preprint arXiv:2208.10442}, 2022.

\bibitem[Wei et~al.(2022)Wei, Wang, Schuurmans, Bosma, Xia, Chi, Le, Zhou, et~al.]{wei2022chain}
Jason Wei, Xuezhi Wang, Dale Schuurmans, Maarten Bosma, Fei Xia, Ed Chi, Quoc~V Le, Denny Zhou, et~al.
\newblock Chain-of-thought prompting elicits reasoning in large language models.
\newblock \emph{Advances in Neural Information Processing Systems}, 35:\penalty0 24824--24837, 2022.

\bibitem[Wu et~al.(2023)Wu, Bansal, Zhang, Wu, Zhang, Zhu, Li, Jiang, Zhang, and Wang]{wu2023autogen}
Qingyun Wu, Gagan Bansal, Jieyu Zhang, Yiran Wu, Shaokun Zhang, Erkang Zhu, Beibin Li, Li Jiang, Xiaoyun Zhang, and Chi Wang.
\newblock Autogen: Enabling next-gen llm applications via multi-agent conversation framework.
\newblock \emph{arXiv preprint arXiv:2308.08155}, 2023.

\bibitem[Zhao and Xu(2023)]{zhao2023less}
Shu Zhao and Huijuan Xu.
\newblock Less is more: Toward zero-shot local scene graph generation via foundation models.
\newblock \emph{arXiv preprint arXiv:2310.01356}, 2023.

\end{thebibliography}
}


\end{document}